\pdfoutput=1
\documentclass[11pt]{article}

\usepackage[final]{acl}

\usepackage{times}
\usepackage{latexsym}

\usepackage[T1]{fontenc}
\usepackage[utf8]{inputenc}

\usepackage{microtype}

\usepackage{inconsolata}

\usepackage{graphicx}
\usepackage{amsmath}
\usepackage{algorithm}
\usepackage{booktabs}
\usepackage{multirow}
\usepackage[noend]{algorithmic}
\newtheorem{definition}{Definition}
\usepackage{enumitem}

\newcommand{\llama}{\includegraphics[height=1em]{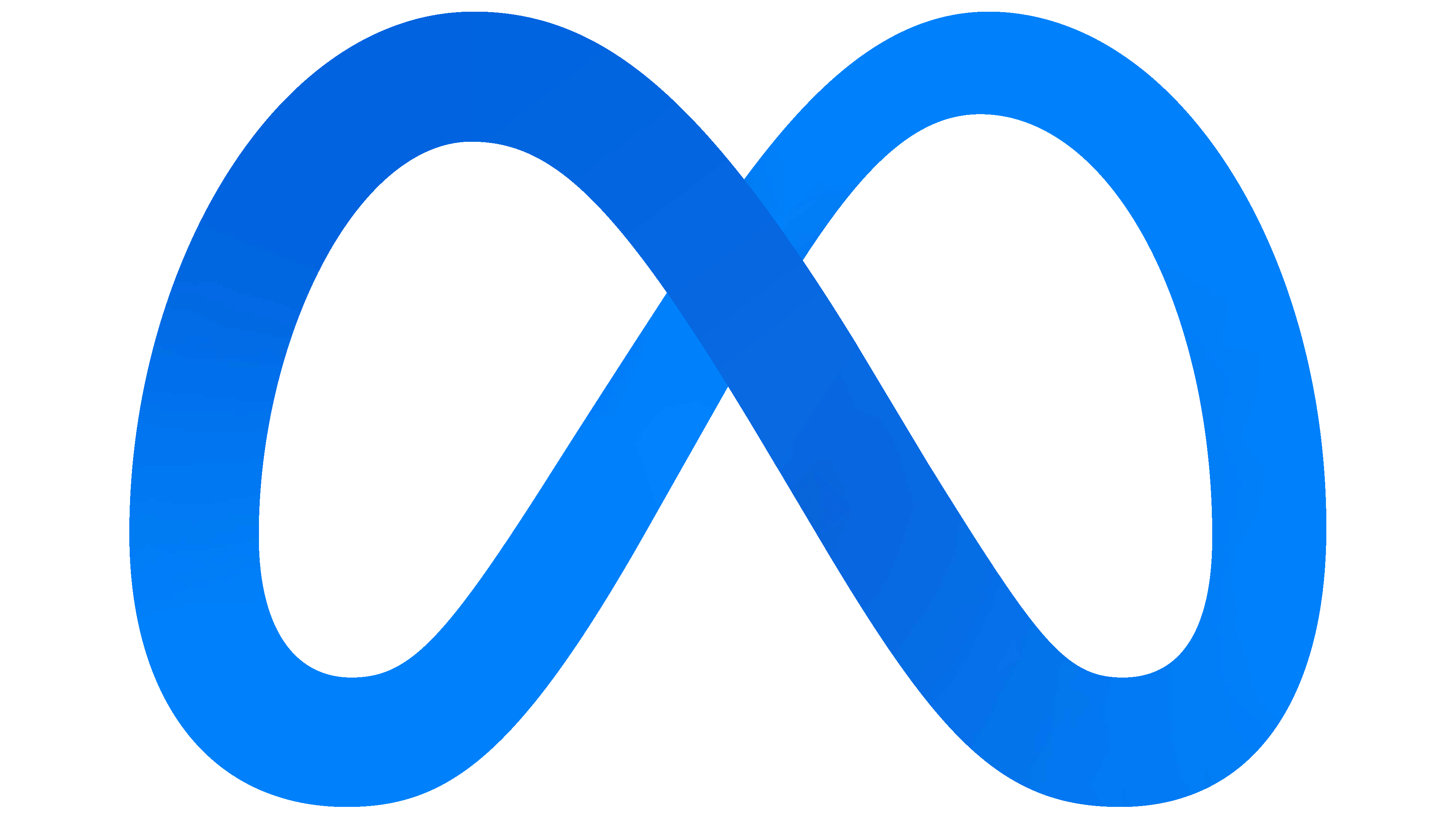}}
\newcommand{\gpt}{\includegraphics[height=1em]{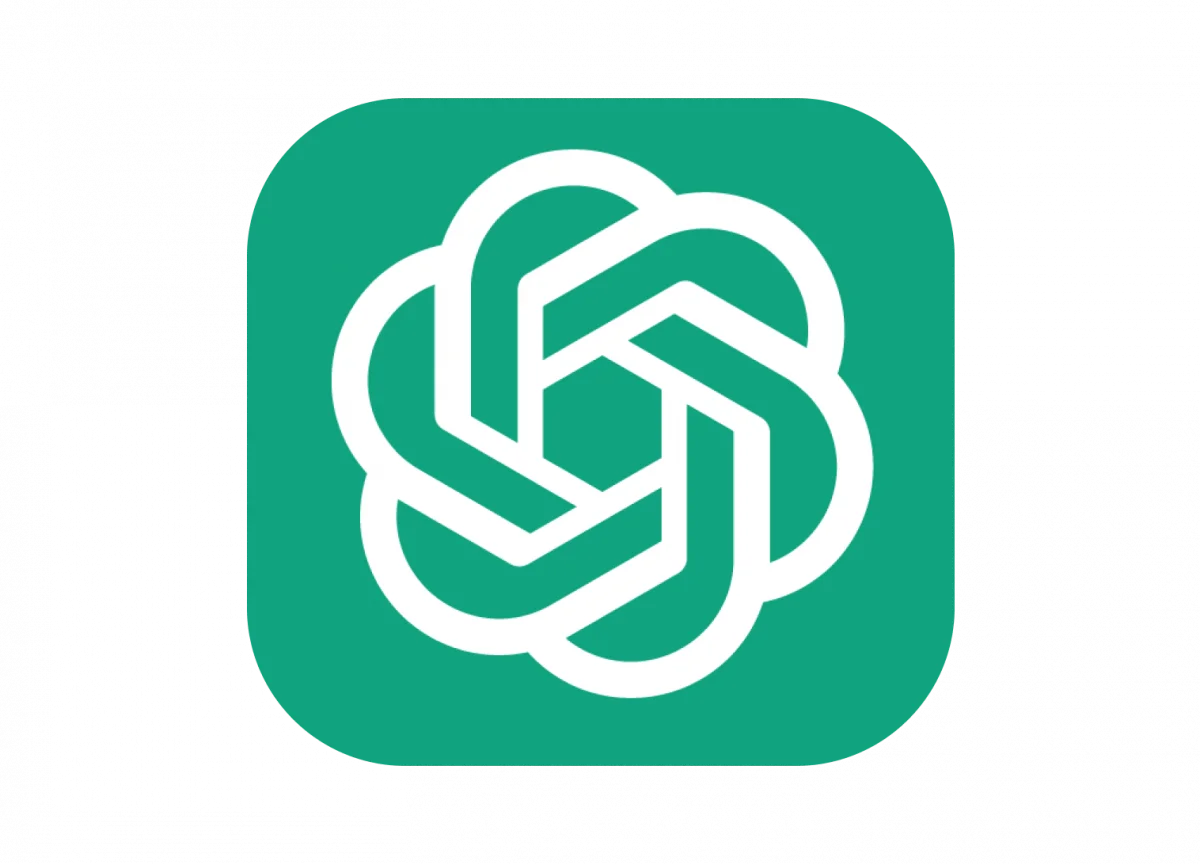}}

\title{TaxoAdapt: Aligning LLM-Based Multidimensional Taxonomy Construction to Evolving Research Corpora}

\author{Priyanka Kargupta$^{\clubsuit}$ \quad Nan Zhang$^\diamondsuit$ \quad Yunyi Zhang$^\clubsuit$ \quad Rui Zhang$^\diamondsuit$\\ {\bf Prasenjit Mitra$^{\diamondsuit}$ \quad Jiawei Han$^\clubsuit$}\\
$^{\clubsuit}$University of Illinois at Urbana-Champaign \quad $^\diamondsuit$The Pennsylvania State University \\
\texttt{\{pk36,yzhan238,hanj\}@illinois.edu}\\ \texttt{\{njz5124,rmz5227,pmitra\}@psu.edu}
}

\begin{document}
\maketitle
\begin{abstract}
The rapid evolution of scientific fields introduces challenges in organizing and retrieving scientific literature. While expert-curated taxonomies have traditionally addressed this need, the process is time-consuming and expensive. Furthermore, recent automatic taxonomy construction methods either (1) over-rely on a specific corpus, sacrificing generalizability, or (2) depend heavily on the general knowledge of large language models (LLMs) contained within their pre-training datasets, often overlooking the dynamic nature of evolving scientific domains. Additionally, these approaches fail to account for the multi-faceted nature of scientific literature, where a single research paper may contribute to multiple \textit{dimensions} (e.g., methodology, new tasks, evaluation metrics, benchmarks). To address these gaps, we propose \textbf{TaxoAdapt}, a framework that dynamically adapts an LLM-generated taxonomy to a given corpus across multiple dimensions. TaxoAdapt performs iterative hierarchical classification, expanding both the taxonomy width and depth based on corpus' topical distribution. We demonstrate its state-of-the-art performance across a diverse set of computer science conferences over the years to showcase its ability to structure and capture the evolution of scientific fields. As a multidimensional method, TaxoAdapt generates taxonomies that are 26.51\% more granularity-preserving and 50.41\% more coherent than the most competitive baselines judged by LLMs.
\end{abstract}

\section{Introduction}

% introduction figure: (from different eras) show two papers' titles, maybe abstract excerpt -> its multi-dimensional contributions
% show two subtrees being expanded based on concepts
\begin{figure}
    \centering
    \includegraphics[width=1.0\linewidth]{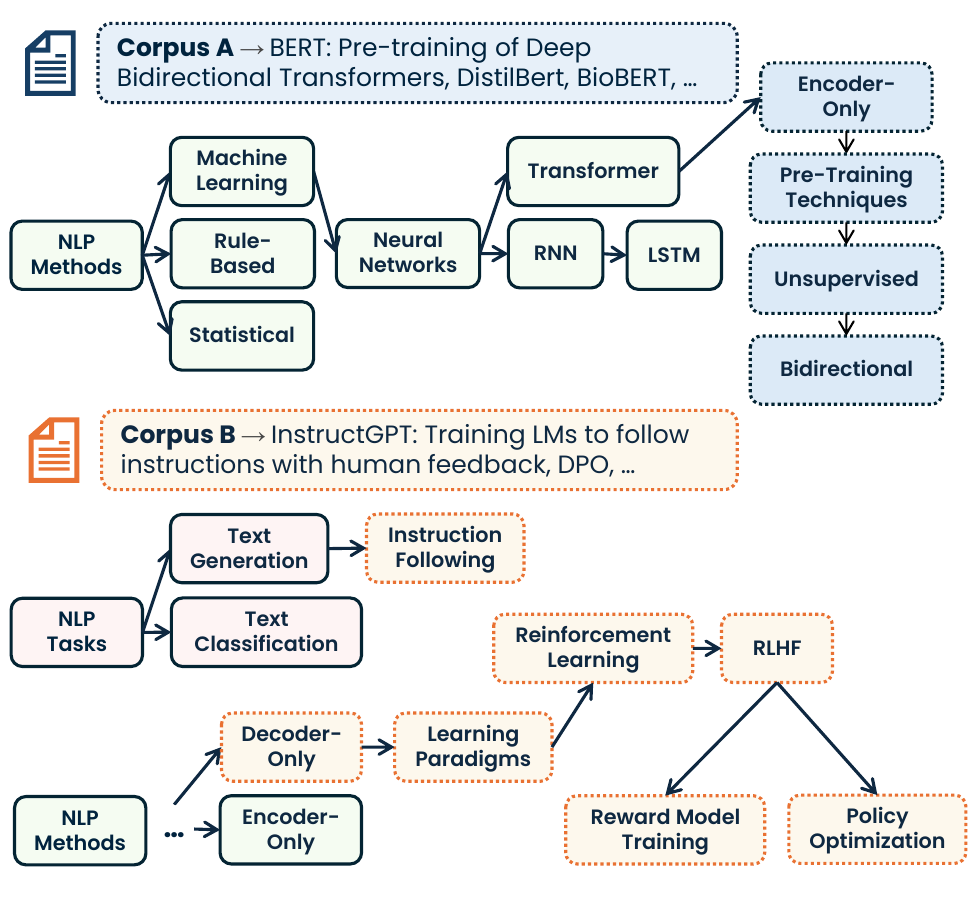}
    \caption{Each paper within a corpus contributes to different dimensions of scientific literature. We show how corpora from different eras of NLP (e.g., BERT-era; RLHF-era) can influence their respective dimension-specific taxonomies (we highlight certain subtrees).}
    \label{fig:example_taxo}
\end{figure}

\par Driven by increased research interest and accessibility, the rapid proliferation of scientific literature and subsequent creation of new branches of knowledge (e.g., the rise of generative models in the last five years) has made organizing and retrieving domain-specific knowledge increasingly challenging \cite{bornmann2021growth,aggarwal2022has}. Taxonomies enhance data organization, support search engines, capture semantic relationships, and aid discovery. While expert-curated and crowdsourced taxonomies have traditionally structured topics into hierarchies (e.g., text classification $\rightarrow$ spam detection), manual curation is time-consuming and struggles to keep pace with rapidly evolving fields \cite{bordea2016semeval,jurgens2016semeval}.

\par Prior efforts in automating taxonomy construction (ATC) fall into two categories: \underline{\textit{corpus-driven}} methods that \textit{extract} topics and relationships directly from text, and \underline{\textit{LLM-based}} approaches which \textit{generate} taxonomies based on pre-existing knowledge. While corpus-driven methods effectively capture meaningful, domain-specific topics, they rely on \textit{rigid approaches} that are restricted to only terms within the corpus vocabulary and lack extensive background knowledge, given their pre-LLM origins \cite{liu2012automatic,shen2018hiexpan,shang2020nettaxo,zhang2018taxogen}.
Conversely, LLM-based methods generate large-scale, general-purpose taxonomies but currently lack mechanisms to align them with specialized knowledge, solely relying on their background knowledge of domains and their key topics \cite{chen2023prompting, shen2024unified, zeng2024chain, sun2024large}.

Moreover, as of now, \textit{both approaches} overlook the \textit{multidimensional} nature of scientific literature. 
A research paper may study and/or contribute to multiple aspects of the scientific method (tasks, methods, applications, etc.), based on which we could organize papers differently. When new knowledge emerges, we must adapt existing taxonomies. For example, in Figure~\ref{fig:example_taxo}, InstructGPT \cite{ouyang2022training} introduces both ``Instruction Following'' as a novel NLP \textit{task} and ``Reinforcement Learning with Human Feedback'' (RLHF) as an NLP \textit{method}, highlighting the limitations of uni-dimensional taxonomies. Limiting ATC design to the task dimension is a \textit{critical oversight}--- obscuring the broader, evolving impacts of research. Ultimately, both corpus and LLM-based methods fail to provide a multidimensional view of scientific literature. To address these gaps, we propose \textbf{TaxoAdapt}, a framework that dynamically grounds LLM-based taxonomy construction to scientific corpora across multiple dimensions. TaxoAdapt operates on three core principles:

\par{\textbf{\underline{Knowledge-augmented} expansion leads to specialized, relevant taxonomies.}} State-of-the-art LLMs struggle to accurately model specialized taxonomies 
in domains like computer science \cite{sun2024large}, particularly leaf-level entities. Existing LLM-based methods require pre-defined entity sets or are limited to entity-level context for taxonomy construction \cite{zeng2024chain,chen2023prompting}, critically limiting the degree of domain-specific knowledge which they can exploit. Alternatively, TaxoAdapt leverages document-level reasoning; by using each paper's title and abstract, it identifies \textit{which} dimensions a paper contributes to (e.g., methods, datasets) and \textit{how}. For example, as shown in Figure \ref{fig:example_taxo}, when expanding the ``Transformer'' node under NLP methods, TaxoAdapt selectively analyzes papers centered on Transformer-based architectures (e.g., BERT)-- helping to derive subcategories like ``Encoder-Only''. Unlike mining important entities, this document-grounded approach enhances taxonomic precision by \textit{aligning expansion with corpus knowledge specific to each dimension, layer, and node}.
% \pm{[Here it is not clear how this method is different from the entity-extraction methods that we criticized before. One could, in principle, mine for the most important entities and discover "Encoder-only" and decide that as a class name, etc.]},

% when to expand
% -> (hierarchical cls) they overlook critical insights from corpus-level distribution (important for scientific domain as this signals trends within the field)
\par{\textbf{Hierarchical text classification provides crucial \underline{signals for targeted exploration.}}} Scientific fields evolve rapidly, with new subdomains emerging and existing ones merging or fading \cite{singh2022quantifying}. Figure \ref{fig:example_taxo} illustrates this: Corpus A (2018–2022) emphasizes BERT-like encoders, while Corpus B (2022–present) highlights ``RLHF'' as a training method and ``Instruction Following'' as a key task behind InstructGPT and its successors. LLM-generated taxonomies often overlook such trends, favoring concepts broadly represented within the training data (e.g., high-level tasks like text classification). 
% \pm{Is this true? Or is this only because there is not a volume of text discussing the new concept. Maybe when RLHF came out then a year or two the LLM will not use this as an important category or sub-category but after a while when the volume grows, LLMs will know and possibly create sub-categories on these entities. How is what you are doing fundamentally different? And, is the problem only that the LLMs are a bit stale and old and the newer ones will fix the problem?}
To address this, TaxoAdapt dynamically adapts the taxonomy by employing hierarchical text classification to determine \textit{which nodes should be expanded 
and how}. A node with a high density of papers (e.g., RLHF) indicates further exploration and warrants \textbf{\textit{depth expansion}} (e.g., Reward Model Training, Policy Optimization). Conversely, if a node has many unmapped papers (e.g., if ``Decoder-Only'' did not exist under ``Transformer''), it signals parallel research to existing children (e.g., ``Encoder-Only''), necessitating \textbf{\textit{width expansion}}. Nodes with minimal presence in the corpus (e.g., LSTMs) will consequently not be explored further.

% how to expand (layer-by-layer controls granularity and reduces redundancy)

\par{\textbf{\underline{Taxonomy-aware clustering} enables meaningful expansion.}} Multiple factors determine which entities should be used to expand a given node: (1) maintaining hierarchical, granular relationships (e.g., identify a \textit{dimension-specific child} of ``Transformer'' and a \textit{sibling} of ``Encoder-Only''), (2) prioritizing presence within the corpus, and (3) minimizing redundancy. Recently, LLMs have shown strong entity clustering abilities \cite{viswanathan2023large, zhang2023clusterllm}. Thus, TaxoAdapt utilizes its knowledge of the dimension, layer, and papers mapped to the specific 
node being expanded to determine granularity-consistent candidate entities. It then utilizes this information to \textit{guide the clustering} of the candidate entities, maximizing coverage while minimizing redundancy during expansion. 
% \pm{[How is the coverage maximized? What does it mean to maximize? And how is redundancy minimized? Explain briefly using an example or give a forward pointer to where you explicate this.]},

Overall, \textbf{TaxoAdapt} aligns the multidimensional taxonomy generation (and expansion) process to a corpus. We summarize our contributions below:
\begin{itemize}
\itemsep0em
\item To the best of our knowledge, \textbf{TaxoAdapt} is the \textit{first} framework to ground LLM-based taxonomy construction to a corpus and study this task from multiple dimensions.
\item We propose a novel classification-based expansion and clustering framework for targeted, meaningful corpus exploration.
\item Through quantitative experiments and real-world case studies, we show that TaxoAdapt outperforms baselines in taxonomic coverage, granular-consistency, and adaptability to emerging research trends.
\end{itemize}

\textbf{Reproducibility:} Our dataset and code is available at \url{https://github.com/pkargupta/taxoadapt}.

% We will provide our dataset and source code to facilitate further studies upon paper acceptance.

\begin{figure*}[ht!]
    \centering
    \includegraphics[width=1.0\linewidth]{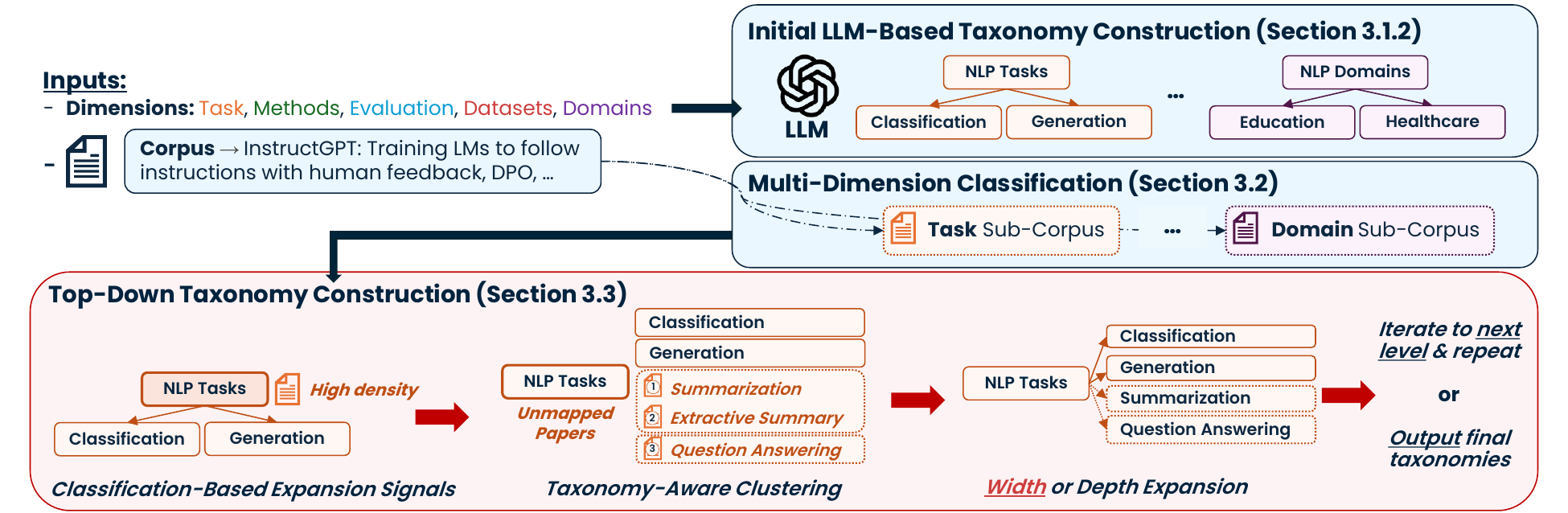}
    \caption{We propose \textsc{\textbf{TaxoAdapt}}, a framework which dynamically constructs a LLM-enhanced, corpus-specific taxonomy using classification-based expansion signals. The diagram demonstrates a \textit{width} expansion example, but the same logic is applied to depth expansion (simply without the additional sibling context).}
    \label{fig:framework}
\end{figure*}

\section{Related Works}
Prior research on taxonomy construction can be broadly categorized into three types: manual, corpus-driven, and LLM-based methods. 
% The latter two represent automatic approaches (ATC).

\paragraph{Manual Curation.} Previous works~\cite{bordea2016semeval,jurgens2016semeval,yang-etal-2013-literature} focused on extracting hand-crafted taxonomies from candidate nodes or designing systems to support the creation of human-assisted taxonomies. These taxonomies involve mostly manual work, making them expensive both during the creation process and for future maintenance, especially given the rapid evolution of scientific fields. Thus, ATC is highly needed.

\paragraph{Corpus-driven Methods.} A line of research~\cite{lu-etal-2024-self,leetaxocom,lee-etal-2022-topic,zhang2018taxogen,huang2020corel} employed clustering to extract entities and their relationships from the corpus, identifying semantically coherent concept terms to complete a given seed taxonomy. Alternatively, NetTaxo~\cite{shang2020nettaxo} leveraged the meta-data of corpus documents as additional signals to construct taxonomies from scratch. Without clustering, HiExpan~\cite{shen2018hiexpan} utilized a relation extraction module to perform depth expansion. Although these approaches maintain a high degree of specificity to the corpus, their lack of LLM usage limits access to broader background knowledge, which is crucial for preserving hierarchical and granular node relationships.

\paragraph{LLM-based Methods.} Many recent works explore the potential of leveraging LLMs for taxonomy expansion or construction. Researchers aimed to answer whether LLMs are good replacement of traditional taxonomies and knowledge graphs, and they found that LLMs still could not capture the highly specialized knowledge of taxonomies and leaf-level entities well~\cite{sun2024large}. In terms of LLM usage, prompting without explicit fine-tuning on any data outperformed fine-tuning-based methods~\cite{chen2023prompting}. TaxoInstruct~\cite{shen2024unified} unified three relevant tasks (entity set expansion, taxonomy expansion, and seed-guided taxonomy construction) by unleashing the instruction-following capabilities of LLMs. Although different iterative prompting approaches~\cite{zeng2024chain,gunn2024creatingfinegrainedentity} have been proposed, there does not exist an LLM-based method that aligns well with the evolving scientific corpus to the best our knowledge. This reinforces our motivation of designing TaxoAdapt.

% Considering the limitations of existing methods, we propose TaxoAdapt to ground LLM capabilities and preserve alignment with the evolving literature as demonstrated in Section~\ref{sec:experimental_results}.
\section{Methodology}
As shown in Figure \ref{fig:framework}, \textsc{TaxoAdapt} aims to align LLM taxonomy generation to a specific corpus, improving adaptability to evolving research corpora. Our framework synergizes both LLM general knowledge and corpus-specific knowledge for automatically constructing more rich and relevant taxonomies.

\subsection{Preliminaries}

\subsubsection{Problem Formulation}
\par We assume that as input, the user provides a topic $t$ (e.g., natural language processing), a set of dimensions $D$ (e.g., tasks, datasets, methods, evaluation metrics), and a scientific corpus $P$. We assume that each paper $p \in P$ is relevant to $t$ and studies at least one $d \in D$. TaxoAdapt aims to output a set of $|D|$ taxonomies $T_{d \in D}$, maximizing the quantity of papers $p \in P$ mapped across all nodes $n_{d} \in T_{d}$. The topic $t$ and dimension $d \in D$ form the root topic $n_0$ of each taxonomy $T_d$ (e.g., ``\textit{natural language processing tasks}''). In order to provide an additional level of flexibility, we define each taxonomy as a directed acyclic graph (DAG) since certain nodes may have two parents (e.g., the scientific question answering (QA) task may be placed under both ``question\_answering'' and ``scientific\_reasoning'').

\subsubsection{Initial LLM-Based Taxonomy Construction}
\label{sec:initial_construction}
\par Recent works \cite{chen2023prompting,sun2024large,zeng2024chain,shen2024unified} have explored leveraging LLMs for taxonomy construction, showing their potential for generating high-level, general-purpose taxonomies (although, these are not guaranteed to be representative of a specific corpus). Given the difficulty of acquiring expert-curated taxonomies across multiple domains and the lack of methods addressing taxonomy construction across multiple dimensions, we utilize an LLM to generate $|D|$ initial single-level taxonomies ($T_{d \in D}$) for \textbf{TaxoAdapt} to expand. This allows us to demonstrate TaxoAdapt's effectiveness while minimizing user input requirements. Nonetheless, this taxonomy can also be replaced by any specific taxonomy which the user desires.
% We provide the prompt in Appendix \ref{app:dag_init}.

\subsubsection{Taxonomy Expansion}
\par Taxonomy expansion involves \textit{both} \textbf{depth} and \textbf{width} expansions of a provided taxonomy, $T_d$. We formally define these below: 
\begin{definition}[\textsc{Depth Expansion}]
\label{def:depth_expansion}
    Expanding a leaf node $n_{i,d} \in T_d$ by identifying a set of child entities $n^i_{j,d} \in N_{d}^i$, which topically falls under $n_{i,d}$ and contains equally granular entities (e.g., $n_{1,d}^i$ and $n_{2,d}^i$ should be equally topically specific). 
    % \rui{notations of $n_{i,d}$ vs $n_{1,d}^i$ is a bit confusing to me.}
\end{definition}
\begin{definition}[\textsc{Width Expansion}]
\label{def:width_expansion}
    Expanding the children of a non-leaf node $n_{i,d}$, where its existing children $n^i_{j,d} \in N_{d}^i$ represent an incomplete set of entities that need to be further completed by additional, unique sibling nodes, $n'^i_d \in {N'}_d^i$. ${N'}_d^i$ and $N_d^i$ are non-overlapping and at the same level of granularity.
    % \rui{notations of $n_{i,d}$ vs $n'^i_d$ is a bit confusing to me. }
\end{definition}

Note that we do not assume a user-provided set of entities for either, which has historically been the case \cite{zeng2024chain, shen2018hiexpan}.

\subsection{Multi-Dimension Classification}
\label{sec:multi_dim_cls}
\par Scientific literature is inherently multifaceted, with individual papers often contributing to multiple aspects of a domain-- such as tasks, methodologies, and datasets. Thus, we must construct a \textit{set} of taxonomies $T_{d \in D}$ that captures the diverse aspects of scientific knowledge. TaxoAdapt seeks to \textit{align} taxonomy $T_d$'s construction with the dimension-specific contributions featured within a corpus.
Thus, we study if and how to minimize the noise present from papers 
that do not make any contributions towards dimension $d$. For example, a paper that only proposes a new text classification dataset,
but still utilizes standard F1-metrics would introduce noise for constructing the ``evaluation method'' taxonomy and consequently, may be omitted. To explore this, we partition the corpus based on the dimensions each paper contributes to before we perform taxonomy expansion.
% \pm{This means that only the first paper that introduced F1-measure is useful and the rest is noise wrt metrics. And, you are ignoring any paper that comes after that as noise. Usually, the number of papers talking about something signifies the importance of a topic and you will miss it. Also, if the first paper calls it the g-measure and then someone renamed it to F-1 and then the community used F-1, will you put both of these in the same bin or will you put them in different bins?}

\par We treat this task as a multi-label classification problem. Recent works have shown that LLMs are successful at fine-grained classification in a multitude of domains \cite{zhang2024teleclass,zhang2024pushing}.
% \jh{Yunyi's paper should be cited as 2025 WWW.} 
Thus, we prompt the LLM to classify the paper $p$, where in-context, we provide the dimension options and their definitions. We define each dimension $d \in D$ with respect to the type of contribution we would expect a paper $p_{i,d}$ to make.
% \pm{d was a subscript and i the superscript in the definitions above. Now, it has flipped. Maybe we should be consistent.}
By default, we assume each paper always falls under the task dimension. We make this assumption because every work has a contribution that is aligned to a specific goal/task. Ultimately, we utilize the output labels for each paper $p \in P$ in order to partition the corpus $P$ into $|D|$ potentially overlapping subsets: $P_d \subseteq P$. We define each of our selected dimensions below:
% (full-length version in Appendix \ref{app: dimensions}):

\begin{itemize}[leftmargin=*]
    \small
    \itemsep=-0cm
    \item \textbf{Task:} We assume all papers are associated with a task.
    \item \textbf{Methodology:} A paper that \textbf{\textit{introduces, explains, or refines a method or approach}}, providing theoretical foundations, implementation details, and empirical evaluations to advance the state-of-the-art or solve specific problems.
    \item \textbf{Datasets:} \textbf{\textit{Introduces a new dataset}}, detailing its creation, structure, and intended use, while providing analysis or benchmarks to demonstrate its relevance and utility. It focuses on advancing research by addressing gaps in existing datasets/performance of SOTA models or enabling new applications in the field. 
    \item \textbf{Evaluation Methods:} A paper that \textbf{\textit{assesses the performance, limitations, or biases of models, methods, or datasets}} using systematic experiments or analyses. It focuses on benchmarking, comparative studies, or proposing new evaluation metrics or frameworks to provide insights and improve understanding in the field. 
    \item \textbf{Real-World Domains:} A paper which demonstrates the use of techniques to solve specific, \textbf{\textit{real-world problems or address specific domain challenges}}. It focuses on practical implementation, impact, and insights gained from applying methods in various contexts. Examples include: product recommendation systems, medical record summarization, etc.
\end{itemize}
% \rui{Not sure if it okay to use smaller font, which may be against policy.}

\subsection{Top-Down Taxonomy Construction}
\label{sec:topdown}
\par An LLM-generated taxonomy may not sufficiently capture all the topics within a corpus, especially in emerging research areas. These areas are underrepresented in the LLMs' general-purpose background knowledge but are highly represented within the input corpus (e.g., the node ``\textit{RLHF}'' in Figure \ref{fig:example_taxo}). Given that domain-specific trends are continually evolving in scientific literature, we must ensure that both the \textit{depth} and \textit{breadth} of the underlying research landscape are accurately represented.

\par To determine which nodes require deeper exploration, we employ hierarchical classification. Adapting an LLM-based text classification model \cite{zhang2024teleclass}, we enrich the taxonomy nodes (e.g., by adding keywords) to support top-down classification from $n_{i,d}$ to $n^i_{j,d}$. Specifically, given a dimension-specific paper $p$ mapped to $n_{i,d}$, we adapt this model to determine whether $p$ (based on its title and abstract) maps to any child node $n_{j,d}^i \in N^i_d$ via multi-label classification using node labels and descriptions. We define $n_{i, d}$'s \textbf{\underline{density}} $\rho(n_{i,d})$ as the \textit{number of papers} $|P_{i,d}|$ mapped to it, leveraging $\rho(n_{i,d})$ to decide whether its children (or lack thereof) should be expanded.

\subsubsection{Depth \& Width Expansion Signals}
\label{sec:signals}

\par When many papers accumulate at a given leaf node $n_{i,d}$, as indicated by a high value of $\rho(n_{i,d})$, it suggests that the topic represented by $n_{i,d}$ is being explored in greater depth within the corpus-- which the current taxonomy does not adequately reflect. Longer taxonomy paths signify popular research topics within the corpus. Figure \ref{fig:example_taxo} illustrates this: the path to ``\textit{bidirectional}'' is significantly deeper than to ``\textit{rule-based}'', reflecting the rise of bidirectional pre-trained language models in Corpus A and the subsequent decline of rule-based methods. In this scenario, if $\rho(n_{i,d}) \geq \delta$ (user-specified threshold), TaxoAdapt performs \textbf{depth expansion} (Definition \ref{def:depth_expansion}) by identifying a set of child entities $N_d^i$ that partition the topic into finer, granularity-consistent subtopics. For instance, as shown in Figure \ref{fig:example_taxo}, if $\rho(\text{``\textit{encoder-only}''}) \geq \delta$, this warrants further decomposition-- such as deepening the path to include ``\textit{pre-training techniques}''-- to capture the ongoing, specialized research in that area.

\par A \textit{complementary} signal is provided by the \textbf{\textit{unmapped} density} $\tilde{\rho}(n_{i,d})$ of a non-leaf node. This arises when a node $n_{i,d}$ has a significant number of papers mapped to it (a high $\rho(n_{i,d})$) that are not allocated to any of its existing child nodes $N_d^i$.
\begin{definition}[\textbf{\textsc{Unmapped Density}}]
Let $P_{i,d}$ denote the set of all papers associated with node $n_{i,d}$, and let $n_{j,d} \in N^i_d$ denote the set of children under node $n_{i,d}$. The unmapped density is then given by:
\begin{equation}
    \label{tag:unmapped_density}
        \tilde{\rho}(n_{i,d}) = \Bigg|P_{i,d} - \bigcup_{j=0}^{|N^i_d|} P_{j,d}\Bigg|
\end{equation}
\end{definition}

If $\tilde{\rho}(n_{i,d})$ exceeds a predefined threshold $\delta$, this indicates that a significant portion of the corpus within $n_{i,d}$ is not adequately represented by its current children. In such cases, TaxoAdapt initiates \textbf{width expansion} by generating additional, non-overlapping sibling nodes $n'^i_{j,d} \in N'^i_d$ to cover the underrepresented research areas. For instance, the ``\textit{decoder-only}'' node in Figure \ref{fig:example_taxo}, where a high $\tilde{\rho}(\text{\textit{``NLP Methods''}})$ signaled that the single ``\textit{encoder-only}'' node did not adequately capture the surge in decoder-only architectures. Once node $n_{i,d}$ is triggered for either depth or width expansion, TaxoAdapt determines the new set of child entities $N'^i_d$ through a pseudo-label clustering procedure (Section \ref{sec:clustering}).

\subsubsection{Taxonomy-Aware Clustering}
\label{sec:clustering}
\par Assuming that node $n_{i,d}$ has been marked for expansion, we must identify a set of child entities ($N'^i_d$ if $n_{i,d}$ is a leaf node, otherwise $N^i_d$) which satisfy the following criteria:
\begin{enumerate}[leftmargin=*]
\itemsep-0.25em
    \item \textit{Maintaining} the hierarchical, granular \textit{relationships} which currently exist within the taxonomy (parent-child and sibling-sibling relationships).
    \item \textit{Maximizing presence} within either the set of unmapped papers $\tilde{\rho}(n_{i,d})$ (width expansion), or $\rho(n_{i,d})$ (depth expansion).
    \item \textit{Minimizing redundancy} between the child entities $N^i_d \cup N'^i_d$.
\end{enumerate}

\par{\textbf{Subtopic Pseudo-Labeling.}} In order to maintain the hierarchical relationships within the taxonomy, we utilize the LLM to generate dimension and granularity-preserving pseudo-labels based on each paper $p_{i,d} \in P_{i,d}$'s title and abstract. We prompt the LLM to determine its dimensional subtopic relative to $n_{i,d}$ as its parent ($n_{i,d}$'s label, dimension, description, and path of ancestors) and $n_{i,d}$'s existing children, if any. 
% We provide the prompt in Appendix \ref{app:pseudo-label}.

\par{\textbf{Subtopic Clustering.}} Given that each paper is represented by its corresponding pseudo-label, clustering these pseudo-labels allows us to \textit{maximize} the number of papers ($\tilde{\rho}(n_{i,d})$ or $\rho(n_{i,d})$) represented. Moreover, effective clustering inherently minimizes redundancy as it aims to produce distinct, non-overlapping sets of papers. We specifically exploit LLM's clustering abilities \cite{viswanathan2023large,zhang2023clusterllm} as this allows us to easily integrate dimension and granularity-specific information into the context and preserve these features within our clusters. Including the same context provided during Subtopic Pseudo-Labeling, in addition to the complete list of paper-subtopic pseudo-labels, we prompt an LLM to determine the primary sub-[dimension] topic clusters (e.g., sub-task, sub-methodology) that would best encompass the list of pseudo-labels, providing a label and description for each cluster. These generated clusters consequently form $N'^i_d$ if $n_{i,d}$ is a leaf node (depth expansion) and otherwise $N^i_d$ (width expansion).

\begin{algorithm}[t!]
\small
    \caption{Top-Down Taxonomy Expansion}
    \label{algorithm: expansion-pseudocode}
    \begin{algorithmic}[1]
        \REQUIRE Topic $t$, Dimension $d \in D$, Corpus $P$, density\_thresh = $\delta$, max\_depth=$l$
        \STATE $T_d \in T = $ initialize\_taxonomy($t, D$) \COMMENT{$T$.depth $= 0$}
        \STATE $P_d \subseteq P \leftarrow $ multi\_dim\_class($t, D$) \COMMENT{Section \ref{sec:multi_dim_cls}}
        \STATE $q$ = queue($\forall \space T_d \in T$)
        \WHILE{$len(q) > 0 \text{ and } T.\text{depth} \leq l$}
            \STATE $n_{i,d} \leftarrow pop(q)$
            \IF{isLeaf($n_{i,d}$)}
                \STATE $n^i_{j,d} \in N^i_d \leftarrow$ expand\_depth($n_{i,d},t$) \COMMENT{Section \ref{sec:clustering}}
                \STATE $q$.append($n_{i,d}$)
            \ELSE 
                \STATE classify\_children($n_{i,d},t,d$) \COMMENT{Section \ref{sec:signals}}
                \IF{$\tilde{\rho}(n_{i,d}) > \delta$}
                    \STATE $n'^i_{j,d} \in N'^i_d \leftarrow$ expand\_width($n_{i,d},t$) \COMMENT{Section \ref{sec:clustering}}
                    \IF{$|N'^i_d| > 0$}
                        \STATE classify\_children($n_{i,d},t,d$)
                    \ENDIF
                    \FOR{$n^i_{j,d} \in N^i_d$}
                        \IF{$n^i_{j,d}$.level $< l$ \algorithmicand $\rho(n^i_{j,d}) > \delta$}
                            \STATE $q$.append($n^i_{j,d}$)
                        \ENDIF
                    \ENDFOR
                \ENDIF
            \ENDIF
        \ENDWHILE
        \RETURN $T$
    \end{algorithmic}
\end{algorithm}

\par We iteratively classify, identify expansion signals, and perform taxonomy-aware clustering level-by-level. We provide the full top-down taxonomy construction algorithm in Algorithm \ref{algorithm: expansion-pseudocode}. Ultimately, this process ends when either no nodes are signaled for expansion or the maximum taxonomy depth is reached---outputting our final $T_d, \space \forall d \in D$.
\section{Experimental Design}
\label{sec:design}
\par We explore \textbf{\textsc{TaxoAdapt}}'s performance using a hybrid of both open (\texttt{Llama-3.1-8B-Instruct}) and closed source (\texttt{GPT-4o-mini}) models. We do this to showcase how we can optimize the cost of the classification and pseudo-labeling steps (both run on \texttt{Llama}) while not needing to sacrifice performance. We discuss our experiment setting details in Appendix \ref{appendix:settings}.

\subsection{Dataset}
\par In order to evaluate \textsc{TaxoAdapt}'s abilities to adapt to different corpora and reflect evolving research topics, we select several conferences spanning different subdomains within computer science. These conferences and their respective sizes are shown in Table \ref{tab:dataset}, where we collect the title and abstract for each paper. We choose to explore our method specifically within computer science such that our dimensions can remain consistent across all conferences: task, methodology, dataset, evaluation methods, and real-world domains. We also include one conference from two different years (e.g., EMNLP'22 and EMNLP'24) in order to showcase how our method reflects the evolution of its respective field.

\begin{table}[h]
\footnotesize
\renewcommand{\arraystretch}{1.1}
\centering
\setlength{\tabcolsep}{5pt}
\caption{Topic $t$ and number of papers (size) per dataset.}
\label{tab:dataset}
\begin{tabular}{l r l}
    \toprule
    \textbf{Conference} & \textbf{Size} & \textbf{Topic $t$} \\
    \midrule
    \textbf{EMNLP 2022} &  828  & \multirow{2}{*}{Natural Language Processing} \\
    \textbf{EMNLP 2024} & 2954  & \\
    \midrule
    \textbf{ICRA 2020} &  1000  & Robotics \\
    \midrule
    \textbf{ICLR 2024}  & 2260  & Deep Learning \\
    \midrule
    \textbf{Total Papers} & 7,042 & \\
    \bottomrule
\end{tabular}
\end{table}

\subsection{Baselines}
\label{sec:baselines}
\par TaxoAdapt aligns LLM-based taxonomy construction to a specialized, multidimensional corpus. Consequently, we choose to compare our method with both \textit{corpus-driven} and \textit{LLM-based} approaches. Note that all LLM-based baselines utilize \texttt{GPT-4o-mini} as their underlying model. We provide detailed information on each baseline in Appendix \ref{appendix: baselines}.

\begin{enumerate}[leftmargin=*]
    \itemsep-0.5em
    \item \textbf{\textit{LLM-Only} $\rightarrow$ Chain-of-Layer \cite{zeng2024chain}:} Given a set of entities, solely relies on an LLM (\textbf{\textit{no corpus}}) to select relevant candidate entities for each taxonomy layer and construct the taxonomy from top to bottom.
    \item \textbf{\textit{LLM + Corpus} $\rightarrow$ Prompting-Based:} An iterative baseline which prompts the LLM to identify relevant papers to the dimension, child nodes, and their corresponding papers.
    \item \textbf{\textit{Corpus-Only} $\rightarrow$ TaxoCom \cite{leetaxocom}:} A corpus-driven, handcrafted taxonomy completion framework that clusters terms from the input corpus to recursively expand a handcrafted seed taxonomy.
    \item \textbf{\textit{No-Dim}} and \textbf{\textit{No-Clustering}} are TaxoAdapt ablations which remove the dimension-specific partitioning and subtopic clustering respectively.
\end{enumerate}
\begin{table*}[ht]
\footnotesize
\centering
\caption{Comparison of models on all datasets, averaged across all dimensions. All values are normalized and scaled by 100. The highest scores for each metric are \textbf{bolded}, and the second-highest scores are marked with a $^\dagger$.}
\label{tab:results}
\renewcommand{\arraystretch}{1.2}

% EMNLP'22 and EMNLP'24
\begin{tabular}{lccccc|ccccc}
\toprule
\multirow{2}{*}{\textbf{Models}} & \multicolumn{5}{c|}{\textbf{EMNLP'22}} & \multicolumn{5}{c}{\textbf{EMNLP'24}} \\
\cmidrule(lr){2-6} \cmidrule(lr){7-11}
 & \textbf{Path} & \textbf{Sib} & \textbf{Dim} & \textbf{Rel} & \textbf{Cover} 
 & \textbf{Path} & \textbf{Sib} & \textbf{Dim} & \textbf{Rel} & \textbf{Cover} \\
\midrule
Chain-of-Layers & 46.87 & 67.67 & 94.61 & 77.65 & 50.54
                & 49.56 & 67.67$^\dagger$ & 92.56$^\dagger$ & 82.13 & 48.66 \\
With-Corpus LLM & 66.14 & 33.93 & 88.82 & 72.87 & 39.35 
                & 49.51 & 29.74 & 83.56 & 84.13$^\dagger$ & 39.20 \\
TaxoCom         & 23.85   & 33.89   & 89.81   & \textbf{91.31}   & \textbf{64.53} 
                & 13.89   & 59.42   & 86.97  & \textbf{95.96}   & 64.81  \\
                \midrule
TaxoAdapt       & 81.09 & \textbf{82.92} & \textbf{100.00} & 82.69$^\dagger$ & 55.81$^\dagger$ 
                & 83.04 & \textbf{77.86} & 98.04 & 88.76$^\dagger$ & 60.29$^\dagger$ \\
- \textit{No Dim}          & \textbf{88.47} & 82.30 & 99.49 & 81.46   & 62.26
                & \textbf{89.98} & 76.97 & \textbf{99.05} & 86.23   & \textbf{66.42}   \\
- \textit{No Clustering}   & 76.45$^\dagger$ & 69.33$^\dagger$ & 98.49$^\dagger$ & 81.63   & 50.38 
                & 65.15$^\dagger$ & 62.15 & 92.31 & 80.22   & 60.80   \\
\bottomrule
\end{tabular}

% ICRA'20 and ICLR'24
\begin{tabular}{lccccc|ccccc}
\toprule
\multirow{2}{*}{\textbf{Models}} & \multicolumn{5}{c|}{\textbf{ICRA'20}} & \multicolumn{5}{c}{\textbf{ICLR'24}} \\
\cmidrule(lr){2-6} \cmidrule(lr){7-11}
 & \textbf{Path} & \textbf{Sib} & \textbf{Dim} & \textbf{Rel} & \textbf{Cover} 
 & \textbf{Path} & \textbf{Sib} & \textbf{Dim} & \textbf{Rel} & \textbf{Cover} \\
\midrule
Chain-of-Layers & 52.92 & 43.46 & 95.06 & 95.00 & 55.96 
                & 40.75 & 43.16 & 95.92 & 69.66 & 48.50 \\
With-Corpus LLM & 74.58 & 32.54 & 97.34 & 94.18 & 45.50 
                & 70.44 & 29.70 & 88.37 & 67.78 & 33.62 \\
TaxoCom         & 43.05   & 54.21   & 99.06$^\dagger$   & 96.28$^\dagger$   & 60.75$^\dagger$
                & 30.00   & 67.00   & 91.27  & \textbf{86.88}   & 56.25$^\dagger$   \\
\midrule
TaxoAdapt       & 86.69 & \textbf{91.59} & \textbf{100.00} & \textbf{97.82} & 52.09 
                & 78.93$^\dagger$ & \textbf{81.47} & 99.62$^\dagger$ & 71.99$^\dagger$ & 53.96 \\
- \textit{No Dim}         & \textbf{91.82} & 89.59$^\dagger$ & \textbf{100.00} & 92.95   & \textbf{67.97} 
                & \textbf{86.32} & 76.45$^\dagger$ & \textbf{100.00} & 69.45   & \textbf{62.54}  \\
- \textit{No Clustering}   & 87.74$^\dagger$ & 85.76 & \textbf{100.00} & 93.97   & 50.86 
                & 65.69 & 67.85 & 93.13 & 68.56   & 54.60  \\
\bottomrule
\end{tabular}
\end{table*}

\subsection{Evaluation Metrics}
\label{sec:evaluation_metrics}
\par We design a thorough automatic evaluation suite using \texttt{GPT-4o} and \texttt{GPT-4o-mini} to determine the quality of our generated taxonomies, using both node-level and taxonomy-level metrics. For each judgment, we ask the LLM to provide additional rationalization (all prompts are in Appendix \ref{appendix:evaluation_prompts}):
\begin{itemize}[leftmargin=*]
    \itemsep-0.25em
    
    \item \textbf{(\textit{Node-Wise}) Path Granularity:} Does the path to node $n_{i,d}$ preserve the hierarchical relationships between its entities (is each child $n_j^i$ more specific than the parent $n_{i,d}$)? Scored 0 or 1 by \texttt{GPT-4o}.
    \item \textbf{(\textit{Level-Wise}) Sibling Coherence:} Determine whether a set of siblings $n_j \in N^i$ of parent node $n_{i,d}$ form a coherent set with the same level of specificity and granularity. Scored from 0 to 1 by \texttt{GPT-4o}.
    \item \textbf{\textit{(Node-Wise}) Dimension Alignment:} Is the node $n_{i,d}$ relevant to the \textit{dimension} $d$ of the root topic $t$? Scored 0 or 1 by \texttt{GPT-4o}.
    \item \textbf{(\textit{Node-Wise}) Paper Relevance:} Is the node $n_{i,d}$ relevant to at least 5\% of the corpus? Scored 0 or 1 per node by \texttt{GPT-4o-mini} (due to longer paper context and thus, cost). Final score is averaged across all nodes.
    \item \textbf{\textit{(Level-Wise)} Coverage:} Given a set of siblings $n_j \in N^i$ of parent node $n_{i,d}$, determine what portion of relevant papers of $n_{i,d}$ are covered by (relevant to) at least one node in the siblings. Scored by \texttt{GPT-4o-mini} (due to longer paper context and thus, cost).
    % \item \textbf{(\textit{Node-Wise}) Paper Relevance:} What proportion of the top $k=10$ papers mapped to this node $n_{i,d}$ are truly relevant to it? Scored by \texttt{GPT-4o-mini} (due to longer paper context and thus, cost).
    % \item \textbf{(\textit{Level-Wise}) Sibling Granularity:} Determine whether a set of siblings $n_j \in N^i$ of parent node $n_{i,d}$ all reflect the same level of specificity. Scored from 1 to 4 (from all different granularities to all the same) by \texttt{GPT-4o}.
    % \item \textbf{\textit{(Level-Wise)} Coverage:} What proportion of papers $p_d \in P_d$ are \textbf{\textit{not}} represented within the taxonomy $T_d$? Scored by \texttt{GPT-4o-mini} (due to longer paper context and thus, cost).
\end{itemize}

\par In addition to this automatic evaluation, we also conduct a supplementary human evaluation for these evaluation metrics. We provide the LLM-human agreement analysis in Appendix \ref{appendix:agreement}. We also provide human evaluation of the subtopic pseudo-labeling and clustering steps (Section \ref{sec:clustering}) in Appendix \ref{appendix:validation}.

\begin{table}[h]
\scriptsize
\centering
\caption{Standard deviation of model performance across all \textbf{datasets \textit{and} dimensions}. 
% TODO: will populate all rows once I get the evaluation results soon\rui{We can remove the Taxocom row if it is empty?}
}
\renewcommand{\arraystretch}{1.2}
\begin{tabular}{lccccc}
\toprule
\textbf{Models} & \textbf{Path} & \textbf{Sib} & \textbf{Dim} & \textbf{Rel} & \textbf{Cover} \\
\midrule
Chain-of-Layers & 0.078 & 0.109 & 0.008 & 0.043 & 0.005 \\
With-Corpus LLM & 0.054 & 0.036 & 0.010 & 0.027 & \textbf{0.004} \\
TaxoCom         & 0.041   & 0.035   & 0.039   & \textbf{0.016}   & 0.022  \\
TaxoAdapt       & \textbf{0.027} & \textbf{0.021} & \textbf{0.007} & 0.043 & 0.015 \\
\bottomrule
\end{tabular}
\label{tab:stdev}
\end{table}

\section{Experimental Results}
\label{sec:experimental_results}

\begin{figure*}
    \centering
    \caption{We show the evolution of NLP Tasks from EMNLP'22 to EMNLP'24. Due to space constraints, we \textbf{\textit{highlight specific subtrees of interest}}, emphasizing nodes which feature commonly-known topical trends within NLP. We also show the number of papers that TaxoAdapt maps to each of the nodes (Section \ref{sec:topdown}) in parentheses. 
    % \rui{Is it possible to make text in the figure larger?}
    }
    \includegraphics[width=1.0\textwidth]{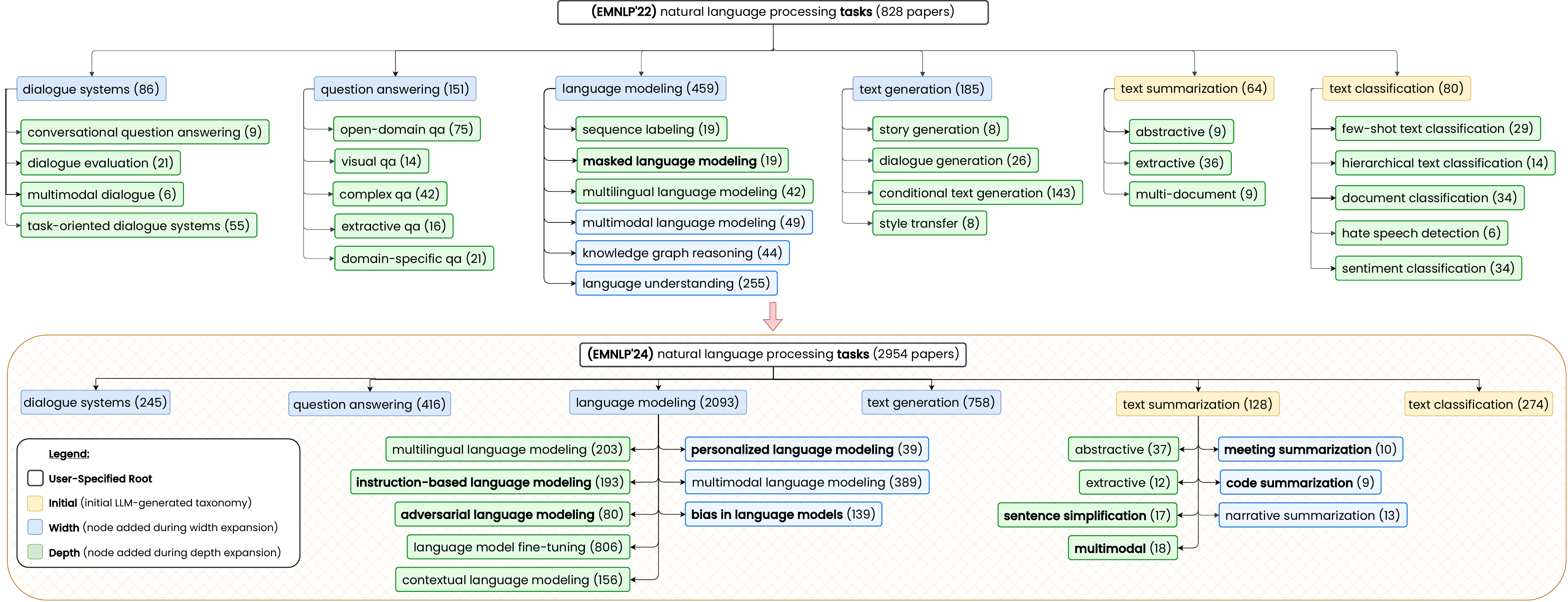}
    \label{fig:comparison}
\end{figure*}

\begin{table*}[!h]
\centering
\small
\begin{tabular}{lccccc|ccccc}
\toprule
\multirow{2}{*}{\textbf{Models}} & \multicolumn{5}{c|}{\textbf{EMNLP'22}} & \multicolumn{5}{c}{\textbf{ICRA'20}} \\
\cmidrule(lr){2-6} \cmidrule(lr){7-11}
& \textbf{Path} & \textbf{Sib} & \textbf{Dim} & \textbf{Rel} & \textbf{Cover} 
& \textbf{Path} & \textbf{Sib} & \textbf{Dim} & \textbf{Rel} & \textbf{Cover} \\
\midrule
Chain-of-Layers \gpt & 46.87 & 67.67 & 94.61 & 77.65 & 50.54
                              & 52.92 & 43.46 & 95.06 & 95.00 & 55.96$^\dagger$ \\
With-Corpus LLM \gpt & 66.14 & 33.93 & 88.82 & 72.87 & 39.35
                              & 74.58 & 32.54 & 97.34 & 94.18 & 45.50 \\
TaxoCom                     & 23.85 & 33.89 & 89.81 & \textbf{91.31} & \textbf{64.53}
                              & 43.05 & 54.21 & 99.06 & 96.28 & \textbf{60.75} \\
\midrule
TaxoAdapt (\llama + \gpt) & \textbf{81.09} & \textbf{82.92} & \textbf{100.00} & 82.69 & 55.81$^\dagger$
                                & 86.69$^\dagger$ & 91.59$^\dagger$ & \textbf{100.00} & 97.82$^\dagger$ & 52.09 \\
TaxoAdapt \llama   & 69.92$^\dagger$ & 74.33$^\dagger$ & 98.70$^\dagger$ & 88.69$^\dagger$ & 51.95
                                & \textbf{92.08} & \textbf{95.11} & \textbf{100.00} & \textbf{98.20} & 49.10 \\
\bottomrule
\end{tabular}
\caption{Comparison of performance across models on EMNLP'22 and ICRA'20 datasets.}
\label{tab:opensource}
\end{table*}

\par{\textbf{Overall Performance \& Analysis.}} Table \ref{tab:results} shows the performance of \textbf{\textsc{TaxoAdapt}} compared with the baselines on a wide variety of node, level, and taxonomy-wise metrics. 
% \rui{not sure which is taxonomy-wise metrics based on discussion from Section 4.3} 
From the results, we can see that TaxoAdapt's taxonomies are \textbf{\textit{$26.51\%$ more granularity-preserving}}, \textbf{\textit{$50.41\%$ more coherent}}, \textbf{\textit{$5.16\%$ more dimension-specific}}, \textbf{\textit{$5.18\%$ more relevant}} to the corpus, and $9.07\%$\textbf{\textit{ more representative}} of the corpus,
compared to the most competitive baseline across all datasets and dimensions. These results indicate that TaxoAdapt is \textit{significantly better} at aligning to a corpus across multiple dimensions, \textit{while still greatly improving the structural integrity of the constructed taxonomies}. Based on our thorough set of experiments, we are able to draw several interesting insights:
\par{\textbf{\textsc{TaxoAdapt} constructs \underline{well-balanced}, cohesive taxonomies.}} We observe that the baselines tend to generate significantly imbalanced taxonomies, where several of the nodes have only a \textit{single child}. Furthermore, each level tends to have an \textit{uncohesive mixture of granularities} (e.g., ``\textit{Sentiment Analysis}'', ``\textit{Emotion Detection}'' as siblings). This is especially the case for TaxoCom, which has a significantly low path granularity while having the highest relevance and coverage score. This is due to it selecting highly coarse-grained nodes (e.g., \textit{NLP tasks $\rightarrow$ significant improvements $\rightarrow$ {closed source, out of domain, text based, \dots}}). In contrast, TaxoAdapt preserves the hierarchical relationships between the topics of taxonomy with cohesive sets of children for each non-leaf node, where the children $n^i_j \in N^i$ of node $n_i$ have high relevance and coverage of $n_i$'s corresponding set of papers $P_i$. Furthermore, each child node $n^i_j$ is relevant to at least $5\%$ of the papers within the corpus $P$, reflected in increased path granularity, sibling cohesiveness, and coverage scores shown in Table \ref{tab:results}. We can attribute these gains to TaxoAdapt's hierarchical classification and taxonomy-aware clustering steps based on the lower performance of ablation, \textit{No Clustering}. We also note that \textit{TaxoAdapt primarily uses \texttt{Llama-3.1-8B} as its backbone model for classification and clustering}, which is a significantly weaker model than the baselines' complete dependence on \texttt{GPT-4o-mini}.

\par{\textbf{\textsc{TaxoAdapt} is \underline{robust} to different research dimensions}.} In addition to each of TaxoAdapt's nodes $n_{i,d} \in T_d$ better reflecting its corresponding dimension (\textbf{\textit{Dim}}), TaxoAdapt exhibits robustness to the different research dimensions. Specifically, Table \ref{tab:stdev} showcases the standard deviation of each model's scores averaged across all dimensions and datasets. We observe that TaxoAdapt features the \textbf{\textit{lowest standard deviations}} across all granularity metrics, while simultaneously scoring the highest for each (Table \ref{tab:results}). We further explore this finding through ablation ``\textit{No-Dim}'', which removes the initial dimension-specific partitioning of the corpus $P$ into $P_{d\in D} \subset P$ (Section \ref{sec:multi_dim_cls}). We observe that partitioning the corpus improves granularity, but also negatively impacts relevance and coverage-- only a narrowed, dimension-specific pool is considered \textit{relevant} for dimension-specific taxonomy construction.

\par{\textbf{\textsc{TaxoAdapt} constructs taxonomies which reflect \underline{evolving research}.} In Figure \ref{fig:comparison}, we demonstrate how TaxoAdapt's taxonomies adapt to corpora from different eras of natural language processing research (EMNLP'22 $\rightarrow$ EMNLP'24). We showcase the task dimension, where due to the rapid increase in EMNLP submissions and accepted papers, features more nodes overall (\textbf{EMNLP'22:} 62 nodes; \textbf{EMNLP'24:} 99 nodes). Furthermore, between the two conference years, we see certain nodes fall in research presence (e.g., \textit{masked language modeling}) and others significantly rise (e.g., \textit{language modeling, instruction-based language models, bias in language models}). We also see certain research trends start to arise as a result of performing \textbf{\textit{width}} expansion based on initially unmapped papers (e.g., \textit{personalized language models}). Overall, Figure \ref{fig:comparison} demonstrates the power of \textbf{\textit{considering classification-based signals}} for \textbf{\textit{knowledge-augmented expansion}}. We include an additional case study of how the taxonomy evolves for the real-world domain dataset using the EMNLP datasets in Appendix \ref{appendix: domain}.

\par{\textbf{Open-Source-Only Performance.}} As mentioned in Section \ref{sec:design}, we optimize the cost of TaxoAdapt by assigning certain tasks to open-source models as opposed to closed-source: (1) \textbf{\texttt{Llama-3.1-8B}:} Dimension classification + hierarchical classification signals + subtopic pseudo-labeling; (2) \textbf{\texttt{GPT-4o-mini}:} Preliminary/initial taxonomy construction (Section \ref{sec:initial_construction}; considered as input into our core framework) and subtopic clustering. Hence, our core framework is built heavily using an open-source model, Llama-3.1. We demonstrate our method’s performance using entirely an open-source model on the EMNLP’22 and ICRA’20 datasets in Table \ref{tab:opensource}.

\par As we can see through TaxoAdapt’s results using \textit{only an 8B open-source model}, its performance across both of the datasets is still \textit{very competitive compared to the GPT-based baselines}, even exceeding our main Llama-GPT variant of TaxoAdapt. This shows that TaxoAdapt is \textbf{very \underline{robust} to different model settings}.

\par{\textbf{Synergizing LLM General and Corpus-Specific Knowledge.}} Appendix \ref{appendix:general} presents a case study and discussion which showcases the power of our corpus-driven, taxonomy-aware framework in synergizing both an LLM's general knowledge and the corpus-specific knowledge for generating more rich and relevant taxonomies.

\par{\textbf{Non-CS Domain Robustness.}} Appendix \ref{appendix:biology} provides an additional quantitative study on TaxoAdapt's performance for a biology dataset--- showcasing that TaxoAdapt still achieves \textbf{\textit{high performance even within more specialized domains}}.

\section{Conclusion}
\par We introduce \textbf{TaxoAdapt}, a novel framework for constructing multidimensional taxonomies aligned with evolving research corpora using LLMs. TaxoAdapt dynamically adapts to corpus-specific trends and research dimensions. Our comprehensive experiments demonstrate that TaxoAdapt significantly outperforms existing methods in granularity preservation, dimensional specificity, and corpus relevance. These results highlight TaxoAdapt's capabilities as a scalable, multidimensional, and dynamically adaptive method for organizing scientific knowledge in rapidly evolving domains.

\section{Limitations}
\par TaxoAdapt relies on LLMs to classify papers into specific dimensions. Although existing works have shown the success of LLMs on fine-grained classification, this classification relies on the parametric knowledge of LLMs, which could be a limitation when LLMs' knowledge becomes outdated. For example, when a dataset paper proposes a new benchmark that has the same (or similar) name as an existing methodology, LLMs might incorrectly assign it to the methodology dimension. However, this is a rare edge case, and TaxoAdapt already generates more dimension-specific taxonomies than baselines as discussed above.

The potential downstream use cases of this taxonomy is to assist with better retrieval \cite{kang-etal-2024-taxonomy} and as a more experimental idea, exploit TaxoAdapt’s coarse and fine-grained signals of where the field is going to inform LLM-based research assistants of both:
\begin{enumerate}
    \item a comprehensive idea of what potential dimension-specific techniques are ``available'' and on-the-rise.
    \item which areas are under-explored for a specific dimension, relative to the research problem they are trying to solve.
\end{enumerate}
\par As these rely on more specialized adaptations of our method (and thus are out of scope), we leave it to future work to explore these potential avenues.

\section{Acknowledgements}
\par This work was supported by the National Science Foundation Graduate Research Fellowship. This research used the DeltaAI advanced computing and data resource, which is supported by the National Science Foundation (award OAC 2320345) and the State of Illinois. DeltaAI is a joint effort of the University of Illinois at Urbana-Champaign and its National Center for Supercomputing Applications.

\bibliography{custom}

\appendix
\appendix
% \section{Appendix}

\section{Experimental Settings}
\label{appendix:settings}
\par We explore \textbf{\textsc{TaxoAdapt}}'s performance using a hybrid of both open (\texttt{Llama-3.1-8B-Instruct}) and closed source (\texttt{GPT-4o-mini}) models. We do this to showcase how we can optimize the cost of the classification and pseudo-labeling steps (both run on \texttt{Llama}) while not needing to sacrifice performance. We construct initial, deterministic single-level taxonomies using \texttt{GPT-4o-mini} (Section \ref{sec:initial_construction}). For all other modules of our framework, we sample from the top 1\% of the tokens and set the temperature to $0.1$. We set the density threshold $\delta$ = 40 papers and the maximum depth $l = 2$. Assuming that the depth of the root is 0 and due to the nature of the task, the size of the taxonomy has the potential to grow exponentially, especially given that the number of child nodes to be inserted is dynamically chosen. Hence, we set the maximum number of levels in the constructed taxonomy to be three ($l=2$). For $\delta$, we choose this by identifying a reasonable number of papers that can fall under a fine-grained category of sufficient interest (avoiding the construction of a very large taxonomy with extremely fine-grained topics). We do not set a dynamic threshold purposefully, so that the expansion can also be influenced by the growth of the field.

\section{Baselines}
\label{appendix: baselines}
\par Our primary motivation for TaxoAdapt is to demonstrate its capabilities of aligning the LLM-based taxonomy construction to a specialized, multidimensional corpus. Consequently, we choose to compare our method with both \textit{corpus-driven} and \textit{LLM-based} approaches. Note that all LLM-based baselines utilize \texttt{GPT-4o-mini} as their underlying model.

\begin{enumerate}[leftmargin=*]
    \item \textbf{\textit{LLM-Only} $\rightarrow$ Chain-of-Layer \cite{zeng2024chain}:} A method which is provided a set of entities and solely relies on an LLM (\textbf{\textit{no corpus}}) to select relevant candidate entities for each taxonomy layer and gradually build the taxonomy from top to bottom. We adapt this method to use an LLM to suggest entities based on the root topic $t$ and dimension $d$.
    \item \textbf{\textit{LLM + Corpus} $\rightarrow$ Prompting-Based:} Given that no methods currently exist which guide LLM taxonomy construction based on a corpus, we design our own prompting-based baseline. Specifically, we conduct an iterative process, where we first ask the LLM to identify relevant papers to the dimension, relevant child nodes, and their corresponding papers. We continue this process until the maximum depth is reached.
    \item \textbf{\textit{Corpus-Only} $\rightarrow$ TaxoCom \cite{leetaxocom}:} A corpus-driven taxonomy completion framework that clusters terms from the input corpus to recursively expand a handcrafted seed taxonomy. We use the same single-level taxonomy from Section \ref{sec:initial_construction} as the seed input, but modify the label names to similar concepts if they do not already exist within the corpus.
\end{enumerate}

\begin{table}[h!]
\small
\resizebox{\columnwidth}{!}{%
\begin{minipage}{\columnwidth}
\caption{Consensus percentages of path granularity, sibling coherence, dimension alignment, and node-paper relevance between LLMs and the human evaluator.}
\label{tab:agreement}
\begin{tabular}{@{}lccc@{}}
\toprule
\textbf{Granularity} & \textbf{Coherence} & \textbf{Alignment} & \textbf{Relevance} \\ \midrule
0.900       & 0.700     & 0.700     & 0.875     \\ \bottomrule
\end{tabular}
\end{minipage}%
}
\end{table}

\section{LLM-Human Agreement Analysis}
\label{appendix:agreement}
Since our automatic evaluation suite is mainly using \texttt{GPT-4o} and \texttt{GPT-4o-mini}, we conduct a small-scale human evaluation to test the reliability of our metrics. Using EMNLP'24, one human evaluator is responsible for validating the LLMs evaluation output on the task dimension of TaxoAdapt's taxonomy. We show the consensus percentage (the percentage of cases where both the LLM and the human evaluator agree on an instance) on path granularity, sibling coherence, and dimension alignment metrics as defined in Section~\ref{sec:evaluation_metrics}. For path granularity, we select 30 random paths from TaxoAdapt's taxonomy and let the human evaluator make independent judgment about the hierarchical relationships between entities (scored 0 or 1 by the evaluator). Similarly, we select 10 random sets of siblings with respect to parent nodes for the evaluator to judge sibling coherence (scored 0.67 or 1 by the evaluator for reasonable or strongest coherence), and 30 random nodes are studied about their alignment to the task dimension (scored 0 or 1 by the evaluator). As for (node-wise) paper relevance and (level-wise) coverage metrics, since they are about evaluating node-paper relevance, we randomly select 16 node-paper pairs (8 pairs are considered relevant while the other 8 are considered irrelevant by \texttt{GPT-4o-mini}) for the evaluator to judge relevance in order to validate these two metrics.

Consensus percentage is shown in Table~\ref{tab:agreement}. The agreement percentages between the LLMs and the human evaluator range from 70\% to 90\%, indicating strong overall agreement. Thus, this human evaluation reinforces the validity of our metrics, so we decide to use them as our automatic evaluation metrics.

\section{Human Evaluation on Subtopic Pseudo-Labeling \& Clustering}
\label{appendix:validation}

\begin{table}[!h]
\centering
\small
\caption{Alignment scores for different pseudo-label types.}
\begin{tabular}{lcc}
\toprule
\textbf{Pseudo-Label Type} & \textbf{Dimension} & \textbf{Paper} \\
\midrule
\textbf{Width Expansion} & 0.8  & 0.8  \\
\textbf{Depth Expansion} & 0.85 & 0.75 \\
\bottomrule
\end{tabular}
\label{tab:pseudo_label_alignment}
\end{table}

\begin{table*}[!ht]
\centering
\small
\begin{tabular}{lccccc}
\toprule
\textbf{Biology Papers} & \textbf{Path} & \textbf{Sib} & \textbf{Dim} & \textbf{Rel} & \textbf{Cover} \\
\midrule
\textbf{Chain-of-Layers} \gpt & 52.69 & 62.99 & \textbf{98.67} & 61.50 & \textbf{49.95} \\
\textbf{TaxoAdapt} (\llama + \gpt) & \textbf{91.08} & \textbf{72.81} & \textbf{98.67} & \textbf{68.23} & 39.70 \\
\bottomrule
\end{tabular}
\caption{Performance comparison on Biology Papers dataset.}
\label{tab:biology_papers}
\end{table*}

\par We have performed two human evaluations to demonstrate the validity of subtopic pseudo-labeling and subtopic clustering (Section \ref{sec:clustering}). Specifically, for pseudo-labeling, we define two binary criteria for verifying the LLM-generated pseudo-labels:
\begin{enumerate}
    \item \textbf{Dimension Alignment:} The pseudo-label aligns with the overall dimension of the taxonomy.
    \item \textbf{Paper Alignment:} The pseudo-label aligns with the titles and abstracts of its corresponding papers.
\end{enumerate}

\par We select 20 papers from width-expanded nodes and 20 papers from depth-expanded nodes. Since each paper comes with a pseudo-label, a human evaluator counts how many labels fulfill these criteria. The proportions of pseudo-labels satisfying each criterion are shown in Table \ref{tab:pseudo_label_alignment}.

\par It is clear to see that the vast majority of pseudo-labels are aligned to both their respective dimensions and papers. This demonstrates the \textbf{validity} and \textbf{effectiveness} of prompting LLMs \textbf{to generate pseudo-labels} for preserving granularities of our taxonomy.

\par As for subtopic clustering, each cluster comes with a name, a description, and a list of pseudo-labels. We define two binary evaluation criteria:

\begin{enumerate}
    \item \textbf{Relevance:} A cluster name needs to capture the majority of its pseudo-labels.
    \item \textbf{Coherence:} All the pseudo-labels of a cluster need to make sense within this cluster.
\end{enumerate}

Randomly selecting 20 clusters, our human evaluator counts the number of clusters that fulfill our criteria. The proportions of clusters satisfying each criterion are shown in Table \ref{tab:cluster_quality}:

\begin{table}[h]
\small
\centering
\caption{Evaluation of cluster quality based on name relevance and coherence. Values indicate the proportion of satisfactory clusters.}
\begin{tabular}{lcc}
\toprule
& \textbf{Relevance} & \textbf{Coherence} \\
\midrule
\textbf{Proportion} & 0.95 & 0.7 \\
\bottomrule
\end{tabular}
\label{tab:cluster_quality}
\end{table}

Both proportions indicate the validity of using LLMs to determine topic clusters. We observe that the proportion of coherent clusters is lower than that of cluster name relevance, since we set a stricter requirement for cluster coherence (all pseudo-labels need to align with the cluster name and description).

\section{Case Study on the Role of LLM General Knowledge in Taxonomy Construction}
\label{appendix:general}

\par The underlying motivation of our work is \textit{how do we adapt LLM-based taxonomy construction to a specific corpus}, which allows the process to be \textit{knowledge grounded} and result in a higher-quality taxonomy overall. Hence, while any method utilizing an LLM will benefit from its general knowledge, we show that \textbf{\textit{LLM general knowledge alone is insufficient for our task}}. We demonstrate this by comparing our method with Chain-of-Layers (only uses an LLM) and With-Corpus LLM (both described in Section \ref{sec:baselines} and Appendix \ref{appendix: baselines}), where we achieve significant performance gains across all metrics-- as shown in Table \ref{tab:results}.

\par TaxoAdapt achieves better performance than Chain-of-Layer across all metrics, which indicates that \textbf{solely using LLMs is not sufficient}. We observe that Chain-of-Layer has a very low path granularity score, which demonstrates a poor hierarchical relationship among entities from top to bottom of its taxonomy. A reason is that Chain-of-Layer is not knowledge-grounded and thus cannot understand fine-grained entities. Despite Chain-of-Layer being provided fine-grained entities present within the corpus as input, it still suffers from poor granularity performance (also seen through the qualitative example below). This indicates that its (GPT-4o-mini’s) general knowledge is insufficient for understanding the hierarchical relationships between these fine-grained entities. In contrast, TaxoAdapt significantly outperforms it using a weaker base model (Llama-3.1-8B) and solely being provided the corpus as input.

\begin{itemize}
  \item Language Model Training
  \begin{itemize}
    \item Parameter Sensitivity in Language Models
    \item Retrieval-oriented Language Model Pre-training
    \begin{itemize}
      \item RetroMAE
    \end{itemize}
    \item Efficient Masked Language Model Training
    \begin{itemize}
      \item Efficient Pre-training of Masked Language Model via Concept-based Curriculum Masking
    \end{itemize}
    \item Intent Detection Frameworks
    \begin{itemize}
      \item Multi-Label Intent Detection
    \end{itemize}
    \item Cross-lingual Summarization Datasets
    \begin{itemize}
      \item EUR-Lex-Sum
    \end{itemize}
    \item Scientific Document Representations
    \begin{itemize}
      \item Contrastive Learning
      \item Citation Embeddings
      \item Similarity-based Learning
      \item Scientific Document Representation
    \end{itemize}
    \item Answer Sentence Selection Models
    \begin{itemize}
      \item Pre-training Transformer Models
    \end{itemize}
  \end{itemize}
\end{itemize}

\par Compared with With-Corpus LLM, TaxoAdapt also delivers a significant improvement, indicating that even integrating corpus-specific information with an LLM is insufficient. With-Corpus LLM has a very low sibling coherence score, which demonstrates its inability of forming coherent sets of sibling nodes. Instead, with our taxonomy-aware pseudo-labeling \& clustering (No Clustering ablation in Table \ref{tab:results}), TaxoAdapt outperforms With-Corpus LLM. This showcases the power of our corpus-driven, taxonomy-aware framework.

\section{Non-Computer Science Domains}
\label{appendix:biology}
\par We originally selected computer science-based papers, as the field naturally features large-scale publicly available papers organized at the conference level. However, we show TaxoAdapt’s performance on a dataset of 1,000 biology papers and compare it against the overall, most competitive baseline, Chain-of-Layers (same experimental settings as the main paper). We can see that despite heavily relying on a \textit{small open-source model}, TaxoAdapt features \textbf{significant gains in the majority of metrics}. We note that coverage score is lower, due to Chain-of-Layers generating more coarse-grained nodes throughout the taxonomy (hence their low path granularity score). This shows that TaxoAdapt still achieves \textbf{high performance even within more specialized domains}.

\begin{figure*}
    \caption{NLP Real-World Domains output taxonomy for EMNLP'22.}
    \includegraphics[width=1.0\textwidth]{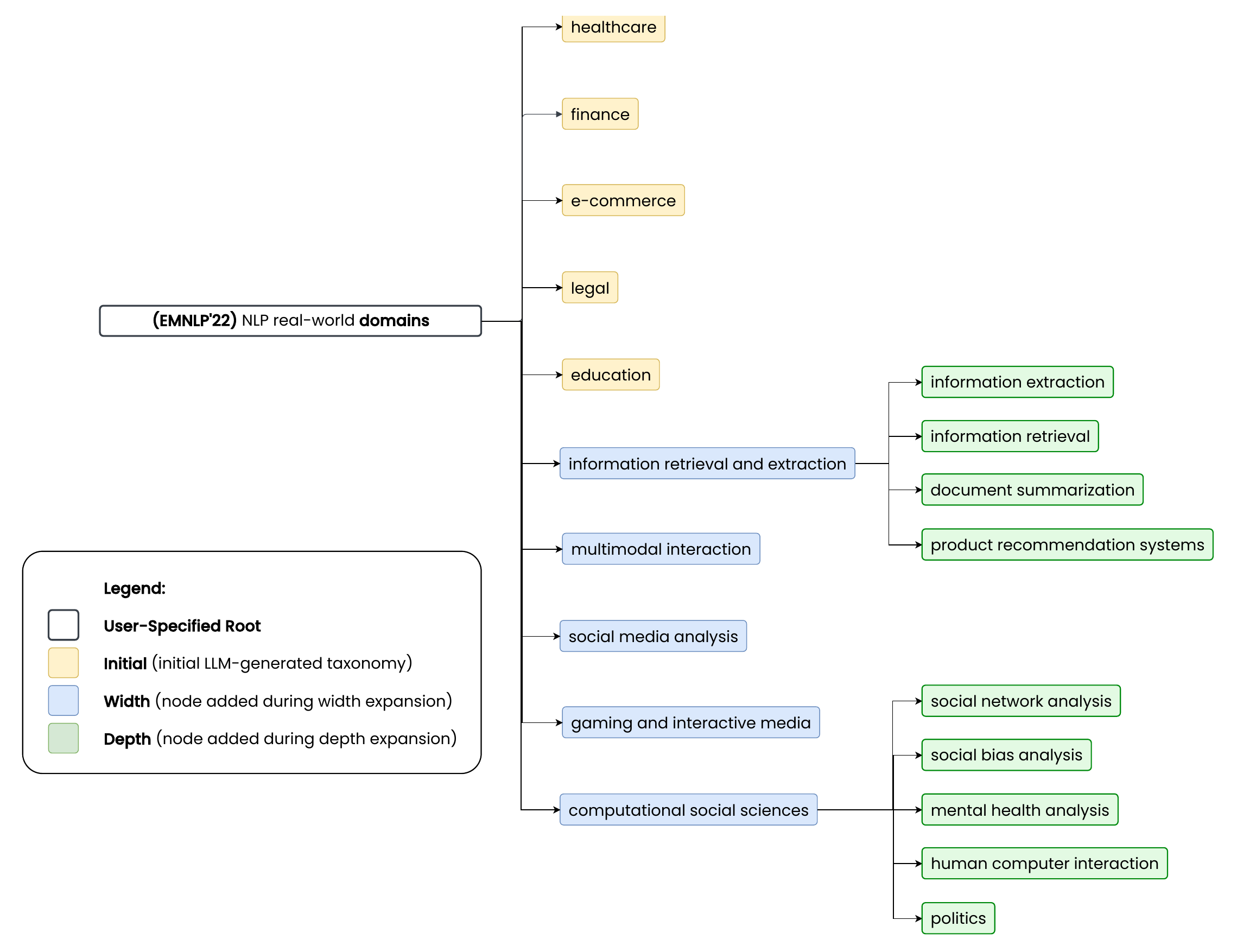}
    \label{fig:domain_22}
\end{figure*}

\begin{figure*}
    \caption{NLP Real-World Domains output taxonomy for EMNLP'24.}
    \includegraphics[width=0.7\textwidth]{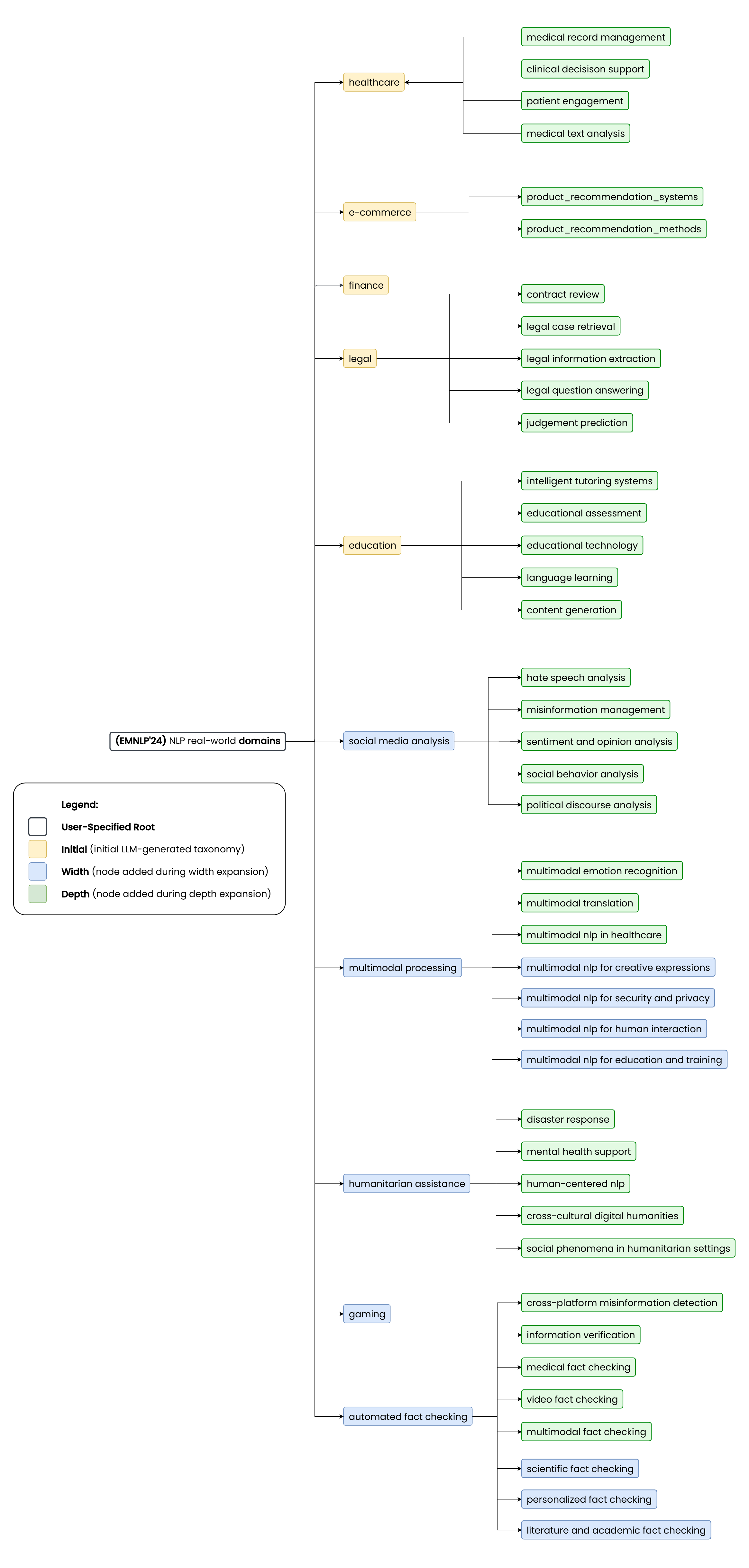}
    \label{fig:domain_24}
\end{figure*}

\section{Case Study on Evolution of NLP Real-World Domains}
\label{appendix: domain}

\par In Figures \ref{fig:domain_22} and \ref{fig:domain_24}, we provide the final outputted taxonomies from TaxoAdapt for the real-world domains dimension of EMNLP'22 and EMNLP'24 respectively. We see that given the rise of large language models, researchers are able to explore the real-world applications of natural language processing in more breadth and depth. This is indicated by the initial LLM-generated nodes (e.g., healthcare, e-commerce) being expanded upon in EMNLP'24 (e.g., medical record management, clinical decision support, patient engagement, etc.). Furthermore, we see more multimodal research as multimodal models have significantly improved. Finally, we see a prominent new node arise in 2024: ``automated fact checking''. This strongly parallels the rise of LLM hallucination as a major public concern. Overall, both case studies on the task and real-world domain dimensions indicate TaxoAdapt's ability to capture evolving research corpora.

\section{LLM Evaluation Prompts}
\label{appendix:evaluation_prompts}
As described in Section~\ref{sec:evaluation_metrics}, we show the LLM prompt that we use to generate evaluation output for computing automatic metrics in Figure~\ref{fig:llm_eval_prompts}.

\begin{figure*}
    \centering
    \includegraphics[width=1.0\textwidth]{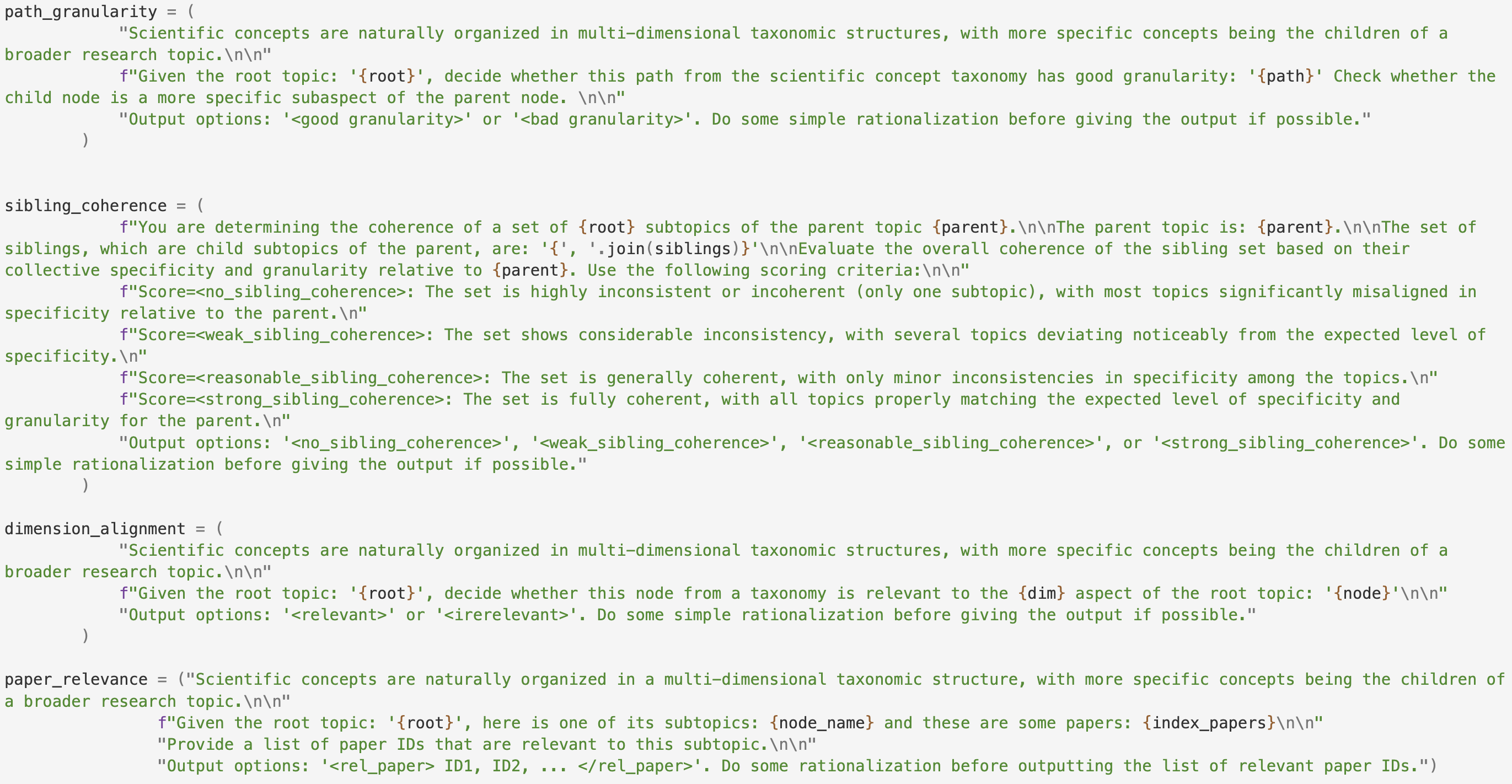}
    \caption{LLM evaluation prompts used to compute path granularity, sibling coherence, dimension alignment, paper relevance, and coverage.}
    \label{fig:llm_eval_prompts}
\end{figure*}

\end{document}